\documentclass[letterpaper]{article} 
\usepackage{aaai25}  
\usepackage{times}  
\usepackage{helvet}  
\usepackage{courier}  
\usepackage[hyphens]{url}  
\usepackage{graphicx} 
\usepackage{natbib}  
\usepackage{caption} 
\usepackage{algorithm}
\usepackage{algorithmic}
\usepackage{amsmath}

\usepackage{ragged2e} 
\usepackage{booktabs,makecell, multirow, tabularx}
\usepackage{newfloat}
\usepackage{listings}
\DeclareCaptionStyle{ruled}{labelfont=normalfont,labelsep=colon,strut=off} 
\floatstyle{ruled}
\newfloat{listing}{tb}{lst}{}
\floatname{listing}{Listing}
\usepackage{colortbl}
\usepackage{float}
\usepackage{placeins}
\title{MoNTA: Accelerating Mixture-of-Experts Training with Network-Traffic-Aware Parallel Optimization}
\author{
    Jingming Guo,
    Yan Liu,
    Yu Meng,
    Zhiwei Tao,
    Banglan Liu,
    Gang Chen,
    Xiang Li
}
\affiliations{
    Shanghai Enflame Technology Co. Ltd, Shanghai, China
}
\usepackage{bibentry}

\begin{document}

\maketitle

\begin{abstract}
The Mixture of Experts (MoE) is an advanced model architecture in the industry that combines multiple specialized expert models from various domains into a single supermodel. This approach enables the model to scale without significantly increasing the computational costs of training and inference, while maximizing model performance. However, current distributed training frameworks do not consider the ultimate optimization of communication, especially for large base models. This paper proposes a network-traffic-aware parallel optimization method that selects the optimal parallel strategy based on the communication volume, and the training cluster's inter-node and intra-node network topologies. Compared to the DeepSpeed, MoNTA achieves an 8x increase in AllToAll communication performance under 8-card tensor parallelism. Compared to the baseline, training a 2x70B model using 16 A800 cards, with an 8K sequence, results in a 13\% overall latency performance improvement. Project Page: https://github.com/EnflameTechnology/DeepSpeed. 

\end{abstract}

%

\section{Introduction}

\begin{figure*}[t]
\centering
\includegraphics[width=0.7\textwidth]{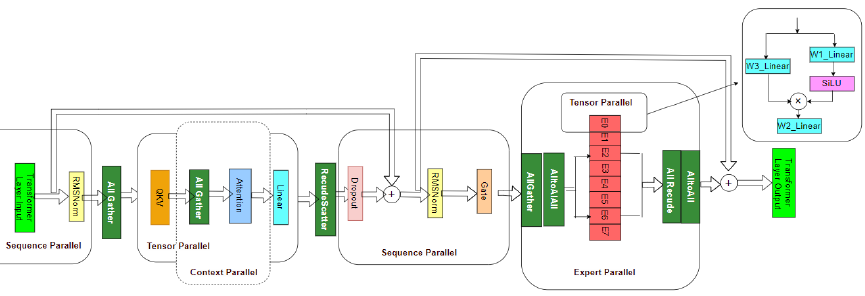} 
\caption{Typical distributed parallelism for MoE models.A common 6D parallel method for MoE models includes Expert Parallel,Context Parallel,Tensor Parallel and Sequence Parallel, where Data Parallel and Pipeline Parallel are not shown.}
\label{fig0}
\end{figure*}

The Mixture of Experts (MoE) model is an ensemble learning approach that combines multiple specialized sub-models or "experts" to enhance the capabilities of large language models without significantly increasing computational cost. Through gating mechanisms, dynamic expert selection is employed to facilitate efficient multitask learning. Due to the substantial communication requirements among the experts in the MoE architecture, overall computational efficiency can be affected. A common 6D parallel method for MoE models within a Transformer structure (excluding DP and PP) is illustrated in Fig.~\ref{fig0}, where experts are typically selected by gating layers, activating only the chosen subset of experts during each forward pass. There is a complex interaction between computational efficiency, communication load, and memory usage. The choice of distributed parallelization strategies will affect these factors and will also be influenced by different hardware configurations.

Currently, distributed training frameworks do not consider the intricate relationship between MoE communication efficiency, communication load, and memory usage, especially for large base models. Existing methods do not address AllToAll communication optimizations for experts under tensor parallelism or leverage inter-node and intra-node communication for parallelism. This paper introduces a network-traffic-aware parallel optimization method that selects the optimal parallel strategy based on the size of the communication volume and the network topology of the training cluster's intra-node and inter-node communications. To the best of our knowledge, this is the first proposal that integrates communication volume, communication efficiency, and network topology in parallel optimization. By exploiting data redundancy in AllToAll communication under tensor parallelism, the AllToAll communication is equivalently transformed into a combination of inter-node AllToAll and intra- node communication. Based on the correspondence between communication volume and communication efficiency, the communication data is divided into different slices to ensure communication efficiency and achieve greater communication overlap. This approach effectively utilizes high-bandwidth intra-node connections to enhance communication efficiency, thereby improving chip compute utilization.

The main contributions of this paper are as follows:
\begin{itemize}
\item We Propose a communication-aware parallel optimization method MoNTA: utilize inter-node and intra-node communication resources, implement inter-node AllToAll and intra-node communication pipelining, establish a performance model for communication volume, communication efficiency, and parallel schemes, achieve MoE AllToAll communication overlap, and improve compute utilization efficiency;
\item We introduce pipelining of intra-node communication and D2D copying to further reduce AllToAll overhead;
\item We analyze communication conflict issues during the training process of the MoE model and provide a communication priority scheme;
\item We propose an expansion method for distributed parallel training cluster of the long context Mixture of Experts model, which generates a distributed parallel expansion strategy for MoE based on cluster resource parameters, model parameters, and context length.
\end{itemize}

Experimental results show that compared to DeepSpeed baseline, MoNTA achieves a maximum performance improvement of approximately 8x in AllToAll communication under 8-card tensor parallelism. Furthermore, compared to the baseline, training a 2x70B model using 16 A800 cards with an 8K sequence results in a 13\% improvement in overall latency performance.

\section{Background and Motivation}

This section introduces the background of MoE structure, expert parallelism, tensor parallelism, and AllToAll communication.
\subsubsection{MoE}

The concept of MoE (Mixture of Experts) first emerged in a paper in 1991. With the emergence and development of sparse-gated MoE, especially when combined with large-scale language models based on Transformers, the continuous expansion of language model capabilities like LLM can be achieved without significantly increasing computational requirements. This year, research related to MoE has shown strong growth, with the successive release of large models such as Mixtral-8x7B~\cite{albert2024mixtral}, Grok-1, DeepSeek-V2~\cite{damai2024deepseekmoe}, and etc. Additionally, there has been a trend of creating large models that combine MoE with other models, such as Jamba~\cite{opher2024jamba}, Samba~\cite{liliang2024samba}, and more.
A typical MoE model consists of two parts: a gating network and a sparse expert layer. The gating layer determines which expert processes a token and is typically composed of linear layers, softmax layers, gating functions (such as TopK), and so on. The sparse expert layer replaces the FFN (Feed-Forward Network) layer in Transformers and is usually composed of multiple FFNs, each representing an independent expert. In Fig. ~\ref{fig1}, there are 8 experts, each with potentially similar or different structures.

\subsubsection{Expert Parallel}

\begin{figure}[t]
\centering
\includegraphics[width=0.47\textwidth]{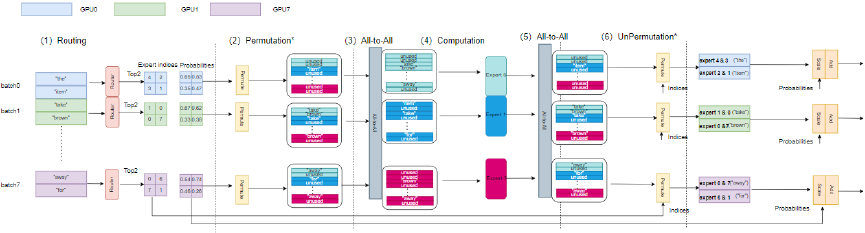} 
\caption{MoE expert parallelism execution process.}
\label{fig1}
\end{figure}

Expert Parallel routes tokens to different experts within Transformer blocks for computation. Each token is routed to a set of different experts, significantly reducing the number of parameters that each token must interact with by skipping certain experts. After communication via GPU-to-GPU AllToAll connections, the experts process the tokens, which must then be sent back to their original GPUs.
In expert parallelism, different experts can be distributed across different compute nodes, enabling data to be processed in parallel, thus improving computational efficiency. Fig. ~\ref{fig1} illustrates the execution flow of expert parallelism in MoE models, typically consisting of several components:
\begin{itemize}
\item Routing: The gating network computes a weight for each expert based on the characteristics of each token, reflecting the importance of that expert in processing the token. On each card, the TopK expert indices and probabilities are selected for each batch data, where K is 2 as shown in Fig. ~\ref{fig1}.
\item Permutation: Reorder the token sequences on each card, grouping data that selects the same expert together to form dispatch encoded data.
\item Dispatch:Using AllToAll communication, distribute the input tokens to the respective experts for processing, so each expert on each card obtains tokens to be processed.
\item Computation:Perform expert computations in parallel on each card.
\item Combine:Utilize AllToAll communication to reconstruct the data, and obtain the original tokens processed by experts.
\item Unpermutation:On each card, restore the original batch order based on the indices from routing. Each token undergoes a scale-add operation using the Top 2 probabilities and the hidden states corresponding to its processing experts, resulting in the final output.
At this point, each token completes the expert computations and proceeds to the next transformer block.
\end{itemize}

\begin{figure}[t]
\centering
\includegraphics[width=0.47\textwidth]{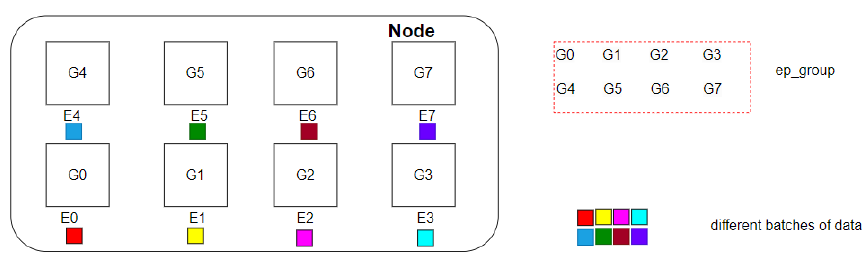} 
\caption{Combination of expert parallelism and data parallelism.}
\label{fig2}
\end{figure}

Expert parallelism can be combined with other forms of parallelism such as tensor parallelism, sequence parallelism, data parallelism, etc., to achieve efficient model training in large-scale cluster systems.
In Fig. ~\ref{fig2}, Non-MOE employs data parallelism, while MOE utilizes expert parallelism, corresponding to the execution flow shown in Fig. ~\ref{fig2}, typically applicable in cases where the number of expert parameters is relatively small.
In the Non-MoE phase, specifically during the MultiHead Attention stage, $dp\_world\_size = 8$, indicating 8-card data parallelism.During the MoE phase, $ep\_world\_size = 8$ and $dp\_world\_size = 1$, indicating 8-card expert parallelism, with one expert placed on each card, while each card also has the full weights for Attention and Gate. Each card runs different batch data, and the Gate outputs the probabilities of the experts selected for each batch. 
Subsequently,$All\-to\-All$ communication is performed within the $ep\_group$ to send tokens to the corresponding TopK experts for computation. 
After each expert completes its calculations, the results for the respective tokens are returned to the GPUs within the $ep\_group$ using $All\-to\-All$ communication.

\subsubsection{AllToAll Communication}

AllToAll communication plays a crucial role in the training MoE models, and its optimization is essential for improving training efficiency and scaling model size. As the parameters of expert models increase, the time consumed by AllToAll communication may account for over 30\% ~\cite{siddharth2023deepseepted}~\cite{changho2021tutel} of the total time. Researchers have proposed various optimization strategies, including load balancing, computation/communication overlap, and memory usage limits. PipeMoE~\cite{shaohuai2023pipemoe}  utilizes parallelism between expert computation and AllToAll communication, employing a micro chunk approach for hiding latency, and presents the optimal parallel degree for pipelining modeling. 
DeepSpeed is a framework developed by Microsoft specifically designed for large-scale distributed training to improve training efficiency and reduce resource consumption. It supports training large models through various optimization strategies like the ZeRO optimizer, mixed-precision training, and pipeline parallelism. DeepSpeed-TED~\cite{siddharth2023deepseepted} addresses the issue of redundant data transmission in AllToAll communication under tensor parallelism by transforming the first AllToAll after the Router into Drop+AllToAll+AllGather. This approach leverages high-bandwidth intra-node communication to enhance the efficiency of AllToAll communication but has not been merged into the official DeepSpeed release. The official DeepSpeed only supports the two MoE parallelization schemes listed in Table 1. For the cases of Non-MoE TP and MoE EP+TP, we have submitted patch~\footnote{https://github.com/microsoft/DeepSpeed/pull/5626} optimizing AllToAll using DTD.
Building upon this, we further propose the network-traffic-aware parallel optimization method named MoNTA.

\begin{table}[t]
\centering
\resizebox{0.4\textwidth}{!}{
\begin{tabular}{c|c|c}
\hline
\ index & \ parallelism & support schemes\\ \hline
1 & Non-MoE TP $\rightarrow$MoE EP & Drop $\rightarrow$AllToAll $\rightarrow$Expert $\rightarrow$AllToAll $\rightarrow$AllGather\\
2 & Non-MoE TP $\rightarrow$MoE EP+TP & AllToAll $\rightarrow$Expert $\rightarrow$AllToAll\\
\hline
\end{tabular}
}
\caption{DeepSpeed support parallel schemes.}
\label{table}
\end{table}

\section{MoNTA}

\subsection{MoE Memory consumption}
Distributed parallel training strategies are constrained by GPU memory storage, and different parallel methods have varying memory resource usage. We analyze the memory consumption for MoE structured models. For ease of reference, we summarize commonly used symbols in Table 2.

\begin{table}[t]
\centering
\resizebox{0.4\textwidth}{!}{
\begin{tabular}{c|c}
\hline
\ Name  & Description\\ \hline
b & microbatch size \\
s & sequence length \\
h & hidden size     \\
a & head number     \\
l & transformer layer number \\
P1 & Non-MoE parameters \\
P2 & MoE Parametes \\
k & numbers of expert selected by each token \\
d & degree of data parallelism \\
p & degree of pipeline parallelism \\
t & degree of tensor parallelism \\
e & degree of expert parallelism \\
\hline
\end{tabular}
}
\caption{Notations.}
\label{table}
\end{table}

The storage of MoE model weights can be divided into two parts: storage for the Non-MoE module $\psi_{1}$ and storage for the MoE module $\psi_{2}$. Taking the Adam optimizer as an example, weights and gradients use fp16/bf16, while the optimizer's momentum is stored in fp32, along with the fp32 master weight. A combination of data parallelism, tensor parallelism, expert parallelism, and pipeline parallelism is utilized, as shown in Table 3.

\begin{table}[t]
\centering
\resizebox{0.4\textwidth}{!}{
\begin{tabular}{c|c|c}
\hline
\ Configurations & \ Non-MoE Memory(Bytes) & MoE Memory(Bytes)\\ \hline
Baseline & $\frac{16\cdot P1}{p\cdot t}$ & $\frac{16\cdot P2}{p\cdot t\cdot e}$\\
Zero O1 & $\frac{(4+\frac{12}{d})\cdot P1}{p\cdot t}$ & $\frac{(4+\frac{12}{d})\cdot P2}{p\cdot t\cdot e}$\\
Zero O2 & $\frac{(2+\frac{14}{d})\cdot P1}{p\cdot t}$ & $\frac{(2+\frac{14}{d})\cdot P2}{p\cdot t\cdot e}$\\
Zero O3 & $\frac{(\frac{16}{d})\cdot P1}{p\cdot t}$ & $\frac{(\frac{16}{d})\cdot P2}{p\cdot t\cdot e}$\\
\hline
\end{tabular}
}
\caption{Memory Footprint of MoE model weights.}
\label{table}
\end{table}

The storage occupancy on a single card consists of model weight storage and activation storage:
\begin{equation}
Mem_{total}=\psi_{1}+\psi_{2}+Mem_{act},
\end{equation}
\noindent where $\psi_{1}$ and $\psi_{2}$ are storage for the Non-MoE module and MoE module, respectively.$Mem_{act}$ is activation storage.

Megatron-LM~\cite{mohammad2020megatronlm}~\cite{vijay2022reducing} estimated the memory usage of activations in the Transformer architecture. Without parallelism, the storage of activations in MoE structures can be calculated as:
\begin{equation}
Mem_{act}=bshl(13+21k+\frac{5as}{h}),
\end{equation}
\noindent

Activation recomputation can reduce storage pressure and is a commonly used memory optimization technique during the training of large models. As shown in Table 4, the memory usage of activation storage varies under different optimization techniques.

\begin{table}[t]
\centering
\resizebox{0.4\textwidth}{!}{
\begin{tabular}{c|c}
\hline
\ Configurations  & Single GPU activation memory(Bytes)\\ \hline
Without parallelism & $bshl(13+21k+\frac{5as}{h})$ \\
PP+TP+EP & $bshl(5+\frac{5k}{e}+\frac{(8+\frac{16k}{e})}{t}+\frac{5as}{h})$ \\
PP+TP+EP+SP & $bshl(\frac{13+\frac{21k}{e}}{t}+\frac{5as}{h})$     \\
PP+TP+EP+SP+selective activation recomputation & $bshl\frac{(13+\frac{21k}{e})}{t}$     \\
PP+TP+EP+SP+full activation recomputation &  $\frac{2bshl}{t}$\\
\hline
\end{tabular}
}
\caption{Memory Footprint of MoE model activation.}
\label{table}
\end{table}

\subsection{Pipelining}
\paragraph{AllToAll and AllGather Pipelining}

In the existing training process of large MoE models, due to the large size of each expert model and the distribution of different experts across nodes, AllToAll communication usually utilizes inter-node bandwidth.Since inter-node bandwidth is smaller than intra-node bandwidth, communication becomes a significant portion of the overall overhead.
TABLE 5 provides the NVLink bandwidth of different Nvidia chips. The unidirectional bandwidth of InfiniBand is 25GB/s or 50GB/s, with an intra-node to inter-node bandwidth ratio ranging from 8:1 to 18:1. Fully utilizing intra-node bandwidth has become one of the effective approaches to optimize AllToAll communication.
\begin{table}[t]
\centering
\resizebox{0.4\textwidth}{!}{
\begin{tabular}{c|c}
\hline
\ Chips  & Interconnect bandwidth GB/s(Unidirectional) \\ \hline
B100/B200 & 900 \\
H100/H20 & 450 \\
H800/A800 & 200 \\
A100 & 300  \\
\hline
\end{tabular}
}
\caption{The interconnect bandwidth of different Nvidia chips.}
\label{table}
\end{table}

\begin{figure}[t]
\centering
\includegraphics[width=0.47\textwidth]{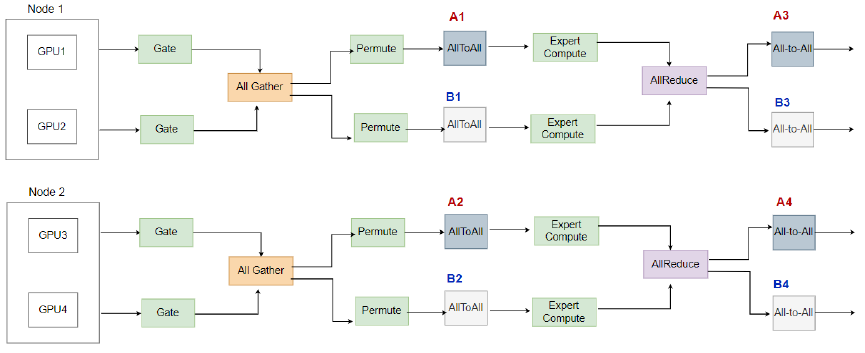} 
\caption{Computation/communication flow of the MoE module under tensor parallelism.}
\label{fig3}
\end{figure}

\begin{figure}[t]
\centering
\includegraphics[width=0.47\textwidth]{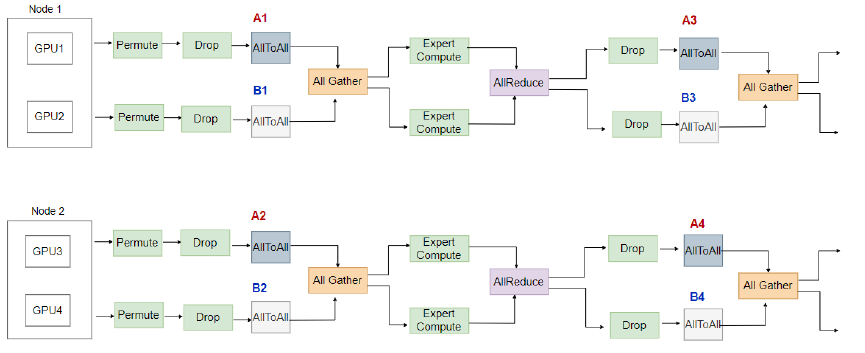} 
\caption{Sequential execution of AllToAll and AllGather.}
\label{fig4}
\end{figure}

Fig. ~\ref{fig3} illustrates the computation/communication flow of the MoE module under tensor parallelism without optimization, as shown in Fig. ~\ref{fig0}. The first AlltoAll communication occurs between GPU1 on Node 1 and GPU3 on Node 2 (A1 and A2), as well as between GPU2 on Node 1 and GPU4 on Node 2 (B1 and B2). AllGather/AllReduce operations occurs between GPU1 and GPU2 within Node 1, and between GPU3 and GPU4 within Node 2. The second AllToAll follows a similar pattern.
DeepSpeed-TED~\cite{siddharth2023deepseepted} leverages the characteristic of identical activations within the tensor parallelism group to reduce redundant data communication during the AllToAll communication. It transforms the first AllToAll after the Router into Drop+AllToAll+AllGather, as shown in Fig. ~\ref{fig4} (with Gather and subsequent AllGather omitted in the figure). By utilizing  high-bandwidth intra-node communication to enhance the overall efficiency of AllToAll communication.

\begin{figure}[t]
\centering
\includegraphics[width=0.47\textwidth]{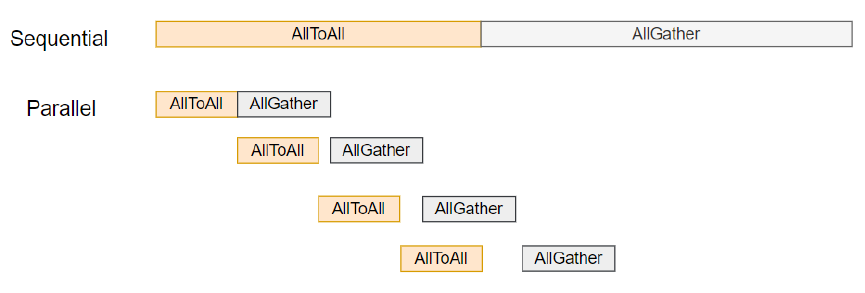} 
\caption{Sequential/Parallel AllToAll and AllGather.}
\label{fig5}
\end{figure}

"Drop" refers to splitting the activations within the tensor parallelism group to ensure that there is no redundant AllToAll data communication within the group. AllToAll communication occurs within the inter-node expert parallel group, with only 1/t of the original communication volume, where each chunk within the tensor parallelism group performs its own AllToAll. AllGather communication takes place within the intra-node tensor parallelism group, gathering the results after AllToAll to achieve equivalence with the original AllToAll. The same method can be applied to the second AllToAll. This paper provides a specific implementation1. However, existing frameworks do not leverage the parallelism of intra-node and inter-node communication, as shown in the sequential method in Fig. ~\ref{fig5}.
To fully utilize the separated hardware architecture resources for inter-node and intra-node communication, this paper proposes a method that parallelizes AllToAll and AllGather, implementing pipelining between inter-node AllToAll and intra-node communication. This approach achieves overlapping of AllToAll communication, avoiding communication resource waiting, and enhances model training efficiency, as illustrated in the parallel method in Fig. ~\ref{fig5}.
The input data for AllToAll communication can be divided into multiple independent chunks, with each chunk executing AllToAll and AllGather operations separately. The AllGather of the current chunk overlaps with the AllToAll of the next chunk. Since AllToAll utilizes inter-node communication resources, typically connected by InfiniBand, while AllGather utilizes intra-node communication resources, usually connected by high-speed NVLink, both are executed on different communication resources in parallel. However,different chunks' AllToAll or AllGather occupy the same resources, leading to their sequential execution. The AllGather of the second chunk needs to wait for the completion of the AllGather of the first chunk, even if the AllToAll of the second chunk has finished.

\begin{figure}[t]
\centering
\includegraphics[width=0.47\textwidth]{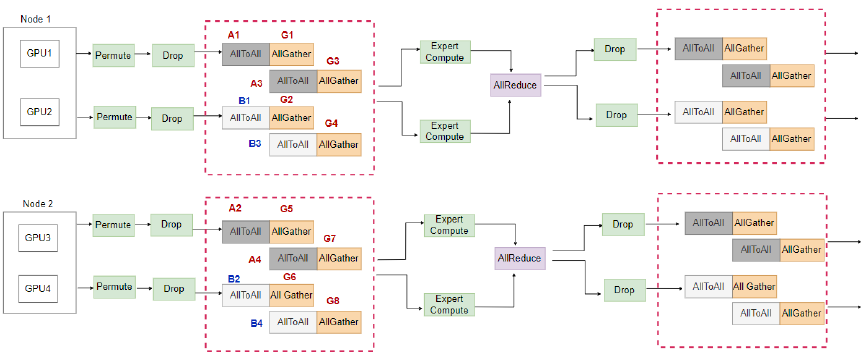} 
\caption{AllToAll and AllGather Pipeling.}
\label{fig6}
\end{figure}

Corresponding to Fig. ~\ref{fig3} and Fig. ~\ref{fig4}, Fig. ~\ref{fig6} illustrates the computation/communication flow of the MoE module under pipelining, where the communication data is divided into two chunks (Gather operator and subsequent AllGather are omitted in the figure).
Step1(AllToAll):The first data chunk performs in inter-node pairwise AllToAll communication, where A1 in Node 1 communicates with A2 in Node 2, and B1 in Node 1 communicates with B2 in Node 2.

Step2(AllGather pipeline with AllToAll): The first data chunk performs intra-node AllGather communication, where G1 in Node 1 communicates with G2, and G5 in Node 2 communicates with G6. At the same time, the second data chunk performs inter-node pairwise AllToAll communication, where A3 in Node 1 communicates with A4 in Node 2, and B3 in Node 1 communicates with B4 in Node 2. This achieves the parallel execution of AllToAll and AllGather as shown in Fig. .

Step3(AllGather): The second data chunk performs  intra-node AllGather communication, where G3 in Node 1 communicates with G4, and G7 in Node 2 communicates with G8.

At this point, the AllToAll communication is complete, and each card proceeds with the Expert computation. The parallel execution of the second AllToAll follows a similar pattern.

\paragraph{AllGather and D2D copy Pipelining}

\begin{figure}[t]
\centering
\includegraphics[width=0.47\textwidth]{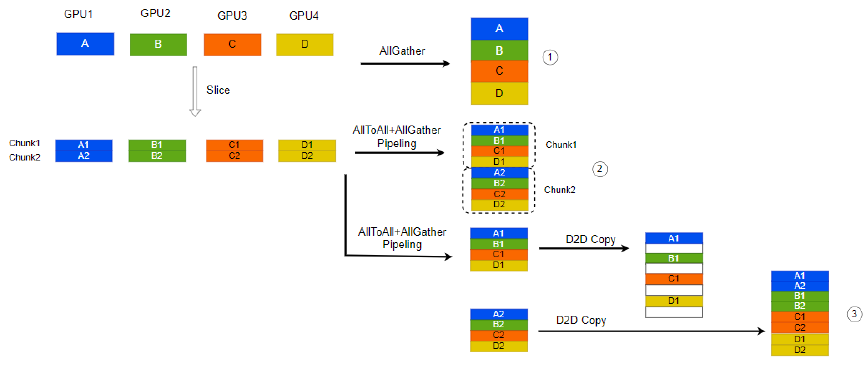} 
\caption{Analysis of data flow for AllToAll and AllGather pipeling.}
\label{fig7}
\end{figure}

\begin{figure}[t]
\centering
\includegraphics[width=0.47\textwidth]{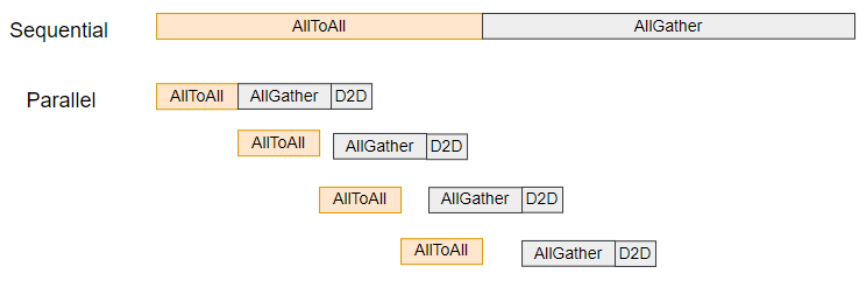} 
\caption{AllGather and D2D copy pipeling.}
\label{fig8}
\end{figure}

We analyze the data flow of parallel execution for AllToAll and AllGather, as shown in Fig. ~\ref{fig7}. For the AllToAll and AllGather in the Sequential mode shown in Fig.~\ref{fig5}, the data arrangement remains the same as in the original unoptimized AllToAll \normalsize{\textcircled{\scriptsize{1}}}\normalsize. In contrast, for the Parallel mode in Fig. ~\ref{fig6} \normalsize{\textcircled{\scriptsize{2}}}\normalsize, the input data is split into chunk1 and chunk2. After pipelining, the final results differ, with data from different chunks arranged in an interleaved manner. To ensure complete data equivalence, Device-to-Device copies are necessary. Once chunk1/chunk2 complete their AllGather operations, they are copied to the corresponding positions based on index offsets \normalsize{\textcircled{\scriptsize{3}}}\normalsize, resulting in the final output that is entirely equivalent to the original data.
The D2D copy operation performed after the AllGather for each chunk can be further optimized to achieve pipelining with AllGather, as shown in Fig. ~\ref{fig8}. The D2D copy of the current chunk and the AllGather of the next chunk are executed in parallel, further overlapping the overall communication time.

\subsection{Network-traffic-aware}
The size of the communication data and the distribution of data between nodes can influence the communication load and efficiency. In theory, the more chunks are split in Fig.~\ref{fig5}, the higher the parallelism of AllToAll and AllGather, and the longer the communication time that can be overlapped. However, the actual execution of communication operators has a fixed overhead~\cite{shriram2001connection}, that is independent of the amount of communication data. Splitting the data into two chunks and executing them sequentially may take longer than completing the operation in a single run.

\begin{figure}[t]
\centering
\includegraphics[width=0.47\textwidth]{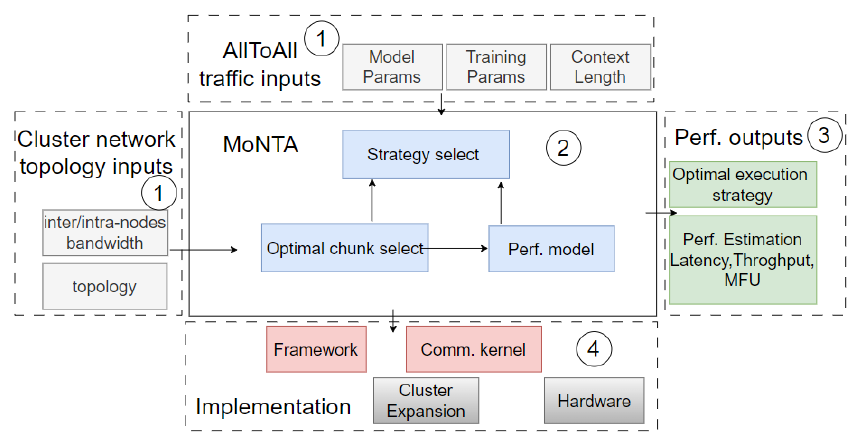} 
\caption{Overview of MoNTA Architecture.}
\label{fig9}
\end{figure}

This paper proposes a network-traffic-aware parallel optimization method called MoNTA, which selects the optimal parallel strategy based on the size of the communication volume and the network topology for intra-node and inter-node connections in the training cluster.
Fig.~\ref{fig9} is a schematic diagram of the MoNTA overview. MoNTA Inputs consist of AllToAll traffic inputs and Cluster network topology inputs \normalsize{\textcircled{\scriptsize{1}}}\normalsize. 
Based on the input information, the optimal chunk size for AllToAll pipelining is searched and selected, and a performance model for various optimization strategies is established to determine the final strategy \normalsize{\textcircled{\scriptsize{2}}}\normalsize. MoNTA outputs the optimal execution strategy while estimating overall latency, throughput, and other performance metrics such as MFU \normalsize{\textcircled{\scriptsize{3}}}\normalsize. 
The accuracy of the performance model is cross-validated through software frameworks, communication operator kernels, and hardware experiments \normalsize{\textcircled{\scriptsize{4}}}\normalsize.  

\textbf{AllToAll traffic inputs:} The total communication volume $I$ for a single AllToAll operation is given by $b*s*h*BPE$, which is related to the model parameters, training parameters, and context length. Here, $b$ represents the microbatch size, $s$ denotes the sequence length, $h$ stands for the hidden size, and $BPE$ signifies the number of bytes per data element.
\begin{equation}
{I}=bsh*BPE,
\end{equation}
\noindent

\textbf{Cluster network topology inputs:} The cluster network topology parameters include theoretical intra-node communication bandwidth, theoretical inter-node communication bandwidth, and the network topology.

\textbf{Optimal chunk select:} When optimizing AllToAll using pipelining, the goal is to select an appropriate chunk size to minimize the overall AllToAll communication time $T$. We assume that the input data for AllToAll communication is divided into $N$ chunks along the sequence dimension, with each chunk containing $s/N$ tokens.
\begin{equation}
{I}=[I_{1},I_{2},\cdots,I_{N}]
\end{equation}
\noindent Here $I_{j}=I[\colon,\frac{(j-1)s}{N}\colon\frac{js}{N},\colon]$,where each chunk communicates independently of the other chunks. We denote the time for AllToAll, AllGather, and D2D copy operations as AA, AG, and D2D, respectively.

\begin{equation}
{AA}=[AA_{1},AA_{2},\cdots,AA_{N}]
\end{equation}
\noindent

\begin{equation}
{AG}=[AG_{1},AG_{2},\cdots,AG_{N}]
\end{equation}
\noindent

\begin{equation}
{D2D}=[D2D_{1},D2D_{2},\cdots,D2D_{N}]
\end{equation}
\noindent

Neglecting effects such as communication network jitter, the theoretical execution time for a single chunk is the same.
\begin{equation}
{AA_{j}}=\frac{I}{N*t}*\frac{e-1}{e}*\frac{1}{B_{1}*r_{1}}
\end{equation}
\noindent

\begin{equation}
{AG_{j}}=\frac{I}{N}*\frac{t-1}{t}*\frac{1}{B_{2}*r_{2}}
\end{equation}
\noindent

\begin{equation}
{D2D_{j}}=\frac{I}{N}*\frac{1}{B_{3}*r_{3}}
\end{equation}
\noindent

Here,$B_{1}$ represents the theoretical bandwidth for AllToAll, $r_{1}$ corresponds to the AllToAll communication efficiency for $I/(N*t)$, which is obtained by referencing the curve that relates communication volume to inter-node communication efficiency based on the value of $I/(N*t)$. 
$B_{2}$ stands for the theoretical bandwidth for AllGather, and $r_{2}$ denotes the AllGather communication efficiency for $I/N$, derived from the curve relating communication volume to intra-node communication efficiency based on the value of $I/N$. The curves relating communication volume to inter-node and intra-node communication efficiency can be pre-tested and plotted. $B_{3}$ is the theoretical memory bandwidth, and $r_{3}$ is the memory bandwidth utilization rate corresponding to $I/N$. 
$r_{1}$,$r_{2}$,$r_{3}$ are variables that change with the variation of data volume.

Algorithm 1 is designed to select the optimal chunk size. The input parameters $b,s,h,BPE$ are related to the AllToAll communication volume, and are associated with formula (3).Input parameters t and e are distributed parallelism parameters related to the selection of parallel strategies.The input parameters $B_1,r_1,B_2,r_2$ are communication-related parameters that can be obtained through pre-testing. The input parameter $I_{minimal}$ represents the minimum communication volume, determining the maximum chunk split N. 
When the size of a single chunk is less than $I_{minimal}$, communication is primarily composed of fixed communication overhead.

\begin{algorithm}[!t]
    \caption{Find Optimal Chunk Size $O_{2}$}
    \hspace*{\algorithmicindent} \textbf{Input}: $b,s,h,BPE,t,e$,$B_{1}$,$r_{1}$,$B_{2}$,$r_{2}$,$B_{3}$,$r_{3}$,$I_{minimal}$ \\
    \hspace*{\algorithmicindent} \textbf{Output}: $N,T$
    \begin{algorithmic}[1]
    \STATE $j=0$
    \STATE $T_{min}=+\infty$
    \STATE $I=bsh*BPE$
    \FOR { $N=1$ to $+\infty$}
        \STATE $I_{AA}=\frac{I}{N*t}$,$I_{AG}=\frac{I}{N}$
        \IF{$I_{AA}<I_{minimal}$ or $I_{AG}<I_{minimal}$ }
           \STATE Break;
        \ENDIF
        \STATE lookup $r_{1}$ according to $I/(N*t)$,lookup $r_{2}$ according to $I/N$ 
        \STATE $AA_{j}=\frac{I}{N*t}*\frac{e-1}{e}*\frac{1}{B_{1}*r_{1}}$
        \STATE $AG_{j}=\frac{I}{N}*\frac{t-1}{t}*\frac{1}{B_{2}*r_{2}}$
        \STATE $D2D_{j}=\frac{I}{N}*\frac{1}{B_{3}*r_{3}}$
        \IF{$AA_{j}<AG_{j}+D2D_{j}$}
           \STATE $T=AA_{j}+(AG_{j}+D2D_{j})*N$
        \ELSE
            \STATE $T=AA_{j}*N+AG_{j}+D2D_{j}$
        \ENDIF
        \IF {$T<T_{min}$}
            \STATE $T=T_{min}$
        \ENDIF
    \ENDFOR
    \STATE Return $N,T$
    \end{algorithmic}
    \end{algorithm}

\textbf{Performance model:} As shown in Table 6, the AllToAll optimization methods are categorized into 3 levels. A performance model is established to achieve network-traffic-aware parallel optimization.

\begin{table}[t]
\centering
\resizebox{0.4\textwidth}{!}{
\begin{tabular}{c|c}
\hline
\ Name  & Method \\ \hline
Baseline & Original AllToAll \\
$O_{1}$ & AllToAll transform into Drop+AllToAll+AllGather \\
$O_{2}$  & AllToAll and AllGather pipelining \\
$O_{3}$ & AllToAll and AllGather pipeling,AllGather and D2D copy pipeling  \\
\hline
\end{tabular}
}
\caption{AllToAll optimization methods.}
\label{table}
\end{table}

Using the $O_{1}$ strategy, the cost model is represented as in Equation (11), where $r_{1}$ is the AllToAll communication efficiency corresponding to $I/t$ and $r_{2}$ is the AllGather communication efficiency corresponding to I.
\begin{equation}
T_{O_{1}}=\frac{I}{t}*\frac{e-1}{e}*\frac{1}{B_{1}*r_{1}} + I*\frac{t-1}{t}*\frac{1}{B_{2}*r_{2}}
\end{equation}
\noindent

Using the $O_{2}$ strategy, the cost model is obtained by the Optimal Chunk Selection module, which searches for the optimal split Nand $T_{O_{2}}$.

Using the $O_{3}$ strategy, by adding pipelining for AllGather and D2D, the cost model also utilizes the Optimal Chunk Selection module to search for the optimal split N and $T_{O_{3}}$ (Algorithm 2). 
Since both AllGather and D2D copy use load/store instructions and are influenced by kernel scheduling, the execution efficiencies $r_{2}$ and $r_{3}$ differ from those of the $O_{2}$ strategy.

\begin{algorithm}[!t]
    \caption{Find Optimal Chunk Size $O_{3}$}
    \hspace*{\algorithmicindent} \textbf{Input}: $b,s,h,BPE,t,e$,$B_{1}$,$r_{1}$,$B_{2}$,$r_{2}$,$B_{3}$,$r_{3}$,$I_{minimal}$ \\
    \hspace*{\algorithmicindent} \textbf{Output}: $N,T$
    \begin{algorithmic}[1]
    \STATE $j=0$
    \STATE $T_{min}=+\infty$
    \STATE $I=bsh*BPE$
    \FOR { $N=1$ to $+\infty$}
        \STATE $I_{AA}=\frac{I}{N*t}$,$I_{AG}=\frac{I}{N}$
        \IF{$I_{AA}<I_{minimal}$ or $I_{AG}<I_{minimal}$ }
           \STATE Break;
        \ENDIF
        \STATE lookup $r_{1}$ according to $I/(N*t)$,lookup $r_{2}$ according to $I/N$ 
        \STATE $AA_{j}=\frac{I}{N*t}*\frac{e-1}{e}*\frac{1}{B_{1}*r_{1}}$
        \STATE $AG_{j}=\frac{I}{N}*\frac{t-1}{t}*\frac{1}{B_{2}*r_{2}}$
        \STATE $D2D_{j}=\frac{I}{N}*\frac{1}{B_{3}*r_{3}}$
        \IF{$AA_{j}<AG_{j}$}
           \STATE $T=AA_{j}+(AG_{j})*N+D2D_{j}$
        \ELSE
            \STATE $T=AA_{j}*N+AG_{j}+D2D_{j}$
        \ENDIF
        \IF {$T<T_{min}$}
            \STATE $T=T_{min}$
        \ENDIF
    \ENDFOR
    \STATE Return $N,T$
    \end{algorithmic}
    \end{algorithm}

\textbf{Strategy Select:} According to the performance model, obtain $T_{O_{1}}$, $T_{O_{2}}$, and $T_{O_{3}}$ ,then select the minimum to determine the optimal strategy S.
\begin{algorithm}[!t]
\caption{Strategy Select}
\hspace*{\algorithmicindent} \textbf{Input}:$T_{O_{1}}$,$T_{O_{2}}$,$T_{O_{3}}$ \\
\hspace*{\algorithmicindent} \textbf{Output}: $S$
\begin{algorithmic}[1]
\STATE Get cost $T_{O_{1}}$,$T_{O_{2}}$,$T_{O_{3}}$
\STATE $T_{min}$=Min($T_{O_{1}}$,$T_{O_{2}}$,$T_{O_{3}}$)
\STATE Get  $T_{min}$ Strategy
\STATE Return S  
\end{algorithmic}
\end{algorithm}

\textbf{Performance outputs:} Based on the output from Strategy Select, determine the optimal execution strategy and estimate the latency, throughput, and Model FLOPs Utilization (MFU) for the training process using this strategy S.

\textbf{Implementations:} Through the framework of optimization strategy implementation and operator kernel implementation, we carry out experiments on hardware clusters, cross-validation and calibrate the performance model. The communication kernel addresses communication conflicts. The implementation also includes cluster expansion strategies for long context MoE training.

\subsection{Communication Conflict}
\begin{figure}[t]
\begin{minipage}[c]{0.7\textwidth}
\flushleft
\includegraphics[width=0.7\textwidth]{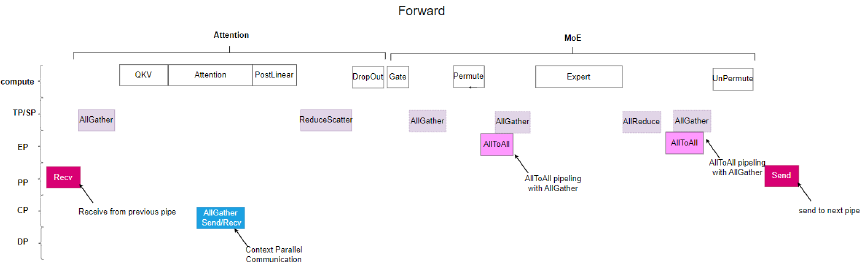} 
\centerline{(a)Forward Computation.}
\end{minipage} \\
\begin{minipage}[c]{0.7\textwidth}
\flushleft
\includegraphics[width=0.7\textwidth]{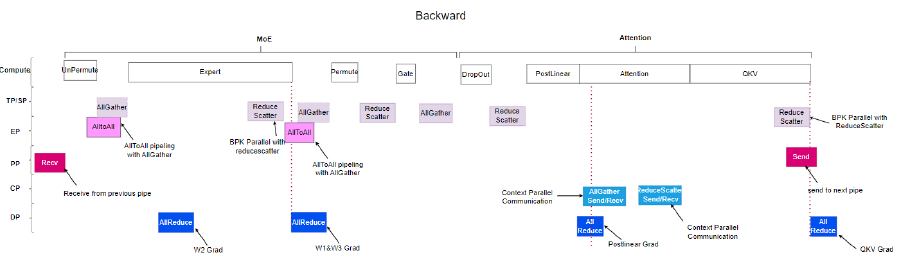} 
\centerline{(b)Backward Computation.}
\end{minipage}
\caption{Communication Conflict}
\label{fig10}
\end{figure}

This section analyzes the communication conflict handling methods in the communication kernel of MoNTA implementations. Fig.~\ref{fig10} illustrates the forward and backward computation and communication timing diagram of the MoE model training process. TP/SP utilize intra-node bandwidth for communication, while EP/PP/CP/DP utilize inter-node bandwidth, which includes southbound scale-out and northbound NIC, among others. EP communication occurs within expert parallel groups, which can be optimized as a combination of inter-node AllToAll communication and intra-node AllGather communication. PP communication involves sending and receiving activations at each pipeline corresponding to the cards. CP communication occurs within context parallel groups, where Attention calculations use RingAttention~\cite{hao2023ring} or AllGather~\cite{abhimanyu2024llama3} operations to pass KVCache, reducing the memory requirement on a single card, typically used for long context training. DP communication involves performing AllReduce on gradients between data parallel group nodes after each batch processing to ensure consistency of model parameters across all nodes.

In Fig. ~\ref{fig10}(a), the vertical axis represents computation, TP/SP, EP, PP, CP, DP, while the horizontal axis represents the timing of execution for the Attention module and MoE module, with each row corresponding to its respective computational/communication operation. If PP uses synchronous communication, during forward computation, there will be no communication conflicts among EP/PP/CP/DP. However, if PP uses asynchronous communication, PP communication during backward may conflict with the forward CP communication's AllGather operation.

In Fig. ~\ref{fig10}(b), when DP uses asynchronous communication, there may be three potential conflicts: (1) Conflict between MoE All2All EP communication and synchronization of W1/W3 gradients; (2) Conflict between CP communication and Postlinear gradient synchronization; (3) Conflict between PP communication and QKV gradient synchronization. If the expert parallelism involves northbound NIC communication, conflicts caused by asynchronous communication between EP and DP/PP can be reduced by setting priorities for different communication primitives: EP>PP>CP>DP, ensuring the efficiency of EP AllToAll communication.

\subsection{Cluster Expansion}
This section analyzes the cluster expansion handling methods for long context MoE training in MoNTA implementations. In the training of the MoE model, it typically involves multiple training steps with Context lengths ranging from 4K to 128K, and even up to 1M tokens. This paper proposes a distributed parallel training extension method for Long Context mixture of expert models. Based on cluster resource parameters, model parameters, and Context length, a strategy for expanding expert model distributed parallelism is generated by combining Expert Parallelism and Context Parallelism.

\begin{figure}[t]
\begin{minipage}[c]{0.7\textwidth}
\flushleft
\includegraphics[width=0.7\textwidth]{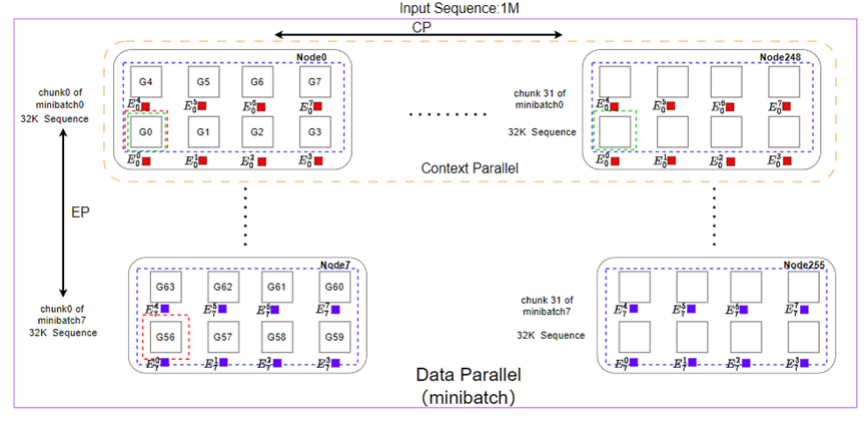} 
\centerline{(a)CP and EP orthogonal.}
\end{minipage} \\
\begin{minipage}[c]{0.7\textwidth}
\flushleft
\includegraphics[width=0.7\textwidth]{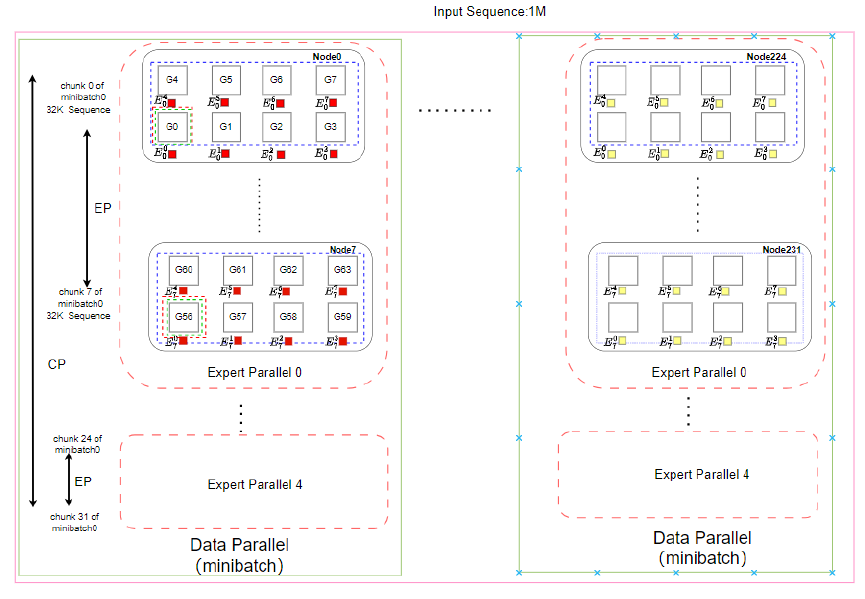} 
\centerline{(b)CP and EP in the same dimension.}
\end{minipage}
\caption{Communication Conflict}
\label{fig11}
\end{figure}

In Fig. ~\ref{fig11}, each node in the current model training cluster consists of 8 GPUs, arranged in the order of [TP/SP, EP, CP, PP, DP]. The tensor parallelism is set to 8, where each of the 8 training cards operates in tensor parallelism. Each training card stores 1/8 of the weights of the Non-MoE modules, as well as 1/8 of the weights of the MoE module. The expert parallelism is set to 8, where the weights of each expert network are split into 8 parts distributed within a node, allowing 8 nodes to store the weights of 8 expert networks. The 8 cards corresponding to the TP positions of 8 nodes form EP Groups.
Both expert parallel and context parallel communications utilize $inter\-node$ communication resources, typically through northbound NIC or southbound $scale-out$. 
In vertical expansion, the tensor parallel groups and expert parallel groups follow the same strategy as horizontal expansion, with CP parallel groups being the point of difference. 
In the vertical expansion strategy, the $CP\_Group$ and $EP\_Group$ are aligned in the same direction.  In vertical CP expansion, the minibatch size of EP Group is 1. 
In horizontal CP expansion, the network topology between EP and CP is orthogonal, requiring a larger switch network to satisfy the communication needs for both EP and CP, with a scale of $cp\_world\_size$ * $ep\_world\_size$.  
In vertical CP expansion, CP and EP are aligned in the same dimension, with the network scale being $max(ep\_world\_size, cp\_world\_size)$.

When the input sequence length is less than what an $EP\_Group$ can handle, i.e. $cp\_world\_size < ep\_world\_size$, and the cluster network switch meets the requirement of $ep\_world\_size$ * $cp\_world\_size$,  horizontal scaling can be employed to enhance single node utilization. As shown in the figure, this can be used to process sequences larger than 32K but smaller than 256K in length.

When the input sequence is greater than or equal to the length that an $EP\_Group$ can handle, i.e., when $cp\_world\_size >= ep\_world\_size$, use vertical expansion. This allows for flexible cluster configuration based on the minimum expansion granularity to meet the GlobalBatch training requirement.

\section{Evaluation}
In this section, we use a 16-GPU A800 cluster with IB 200Gb/s connections between nodes. First, we tested the IB utilization during AllToAll communication with 16 cards under different communication volumes. Then, we establish performance models under different optimization strategies and compare the performance of the AllToAll under these strategies against the DeepSpeed baseline. 
After that, we evaluate the loss convergence of a 2x70B model under different optimization strategies and test the end-to-end model performance. Finally, we discuss the experimental results.

\subsection{Experiment Settings}
\textbf{Hardware platform:} We utilize a 2-node, 16-GPU cluster with nodes interconnected via IB 200Gb/s. Each node consists of 8 A800 SXM4 GPUs with 80GB HBM2, interconnected using NVLink at 400GB/s.

\textbf{Software platform:} The experiments are conducted under a software environment of PyTorch 2.2, CUDA Toolkit 12.3, NCCL 2.19.3, and Ubuntu 22.04, along with Transformers 4.38.1 and DeepSpeed 0.14.2.

\textbf{MoE configurations:} We use a 2x70B MoE (Mixture of Experts) model, with each expert occupying one node. Specific experimental parameters are listed in Table 7. To measure the performance of parallelization schemes under different communication volumes, the sequence length ranges from 4K to 256K.

\begin{table}[t]
\centering
\resizebox{0.4\textwidth}{!}{
\begin{tabular}{c|c}
\hline
\ Parameter  & Value\\ \hline
Sequence Length(s) & 4K~256K \\
$DP\_Size$ & 2 \\
$TP\_Size$ & 8     \\
$EP\_Size$ & 2     \\
Hidden Dim(h) & 8192 \\
\hline
\end{tabular}
}
\caption{Configurations of MoE.}
\label{table}
\end{table}

\subsection{AllToAll performance}
\begin{figure}[t]
\begin{minipage}[c]{0.7\textwidth}
\flushleft
\includegraphics[width=0.7\textwidth]{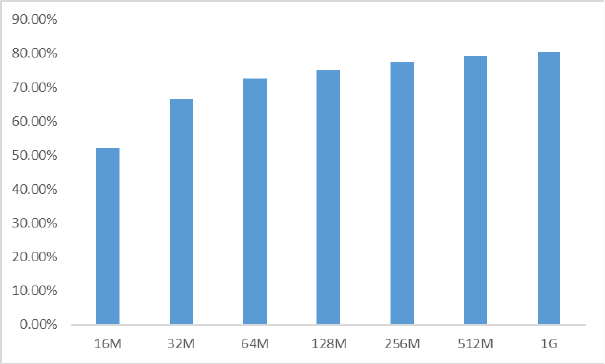} 
\centerline{(a)Efficiency of 8-card AllGather Communication.}
\end{minipage} \\
\begin{minipage}[c]{0.7\textwidth}
\flushleft
\includegraphics[width=0.7\textwidth]{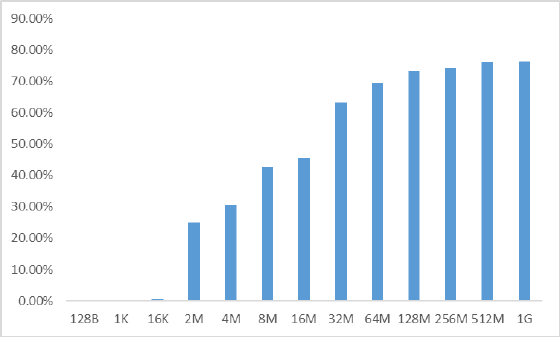} 
\centerline{(b)Efficiency of 16-card AllToAll Communication.}
\end{minipage}
\caption{Communication Efficiency}
\label{fig12}
\end{figure}

\textbf{Communication Efficiency:}We test the communication efficiency of 8-card intra-node AllGather, as shown in Fig.~\ref{fig12}(a). The communication efficiency of the 16-card AllToAll with pxn enabled under varying communication volumes is presented in Fig.~\ref{fig12}(b). For the 2-node 16-card setup, NVLink + IB network cards are used, with the bottleneck being the IB network card. The equivalent bandwidth of the IB network card is calculated using the formula $(S*(N-8))/(N*t)$, where S is the original communication volume $bsh$ , N is the number of cards, and t is the communication time. It can be observed that when the data volume is very small, latency dominates, resulting in nearly 0 bandwidth utilization.

\begin{figure}[t]
\centering
\includegraphics[width=0.47\textwidth]{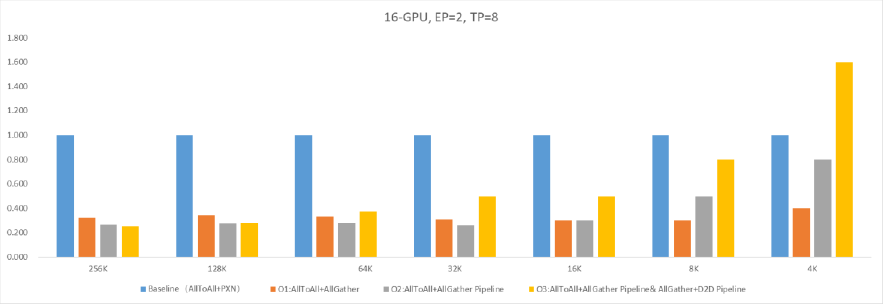} 
\caption{AllToAll performance.}
\label{fig13}
\end{figure}

We establish a performance model under different optimization strategies based on communication efficiency. Based on this, in Fig.~\ref{fig13},  we test the performance of the AllToAll single operator under different communication volumes. It can be observed that as the volume of communication increases, the performance improves more significantly with $O_{1}/O_{2}/O_{3}$ optimizations. When the communication volume decreases, the performance of $O_{1}$ may surpass that of $O_{2}/O_{3}$.
In Fig.~\ref{fig13}, all sequences are split into 4 chunks for both $O_{2}$ and $O_{3}$. When the sequence is 256K, MoNTA selects $O_{3}$. When the sequence is between 32K and 128K, MoNTA chooses $O_{2}$. When the sequence is less than 16K, MoNTA selects $O_{1}$.

\begin{figure}[t]
\centering
\includegraphics[width=0.47\textwidth]{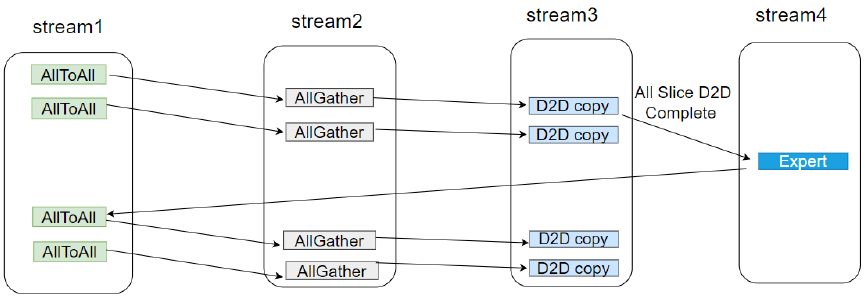} 
\caption{Execute streams.}
\label{fig14}
\end{figure}

\textbf{Implementation:} Establish multiple streams to run different kernels, as shown in Fig.~\ref{fig14}. Stream 1 runs the AllToAll inter-node communication kernel, Stream 2 runs the AllGather intra-node communication kernel, Stream 3 runs the D2D copy kernel, and Stream 4 runs the exert computation kernel.

Assuming the input sequence is divided into two data chunks, within Stream 1, the first two AllToAll operations represent the communication data chunks processed before the expert computation, while the last two AllToAll operations represent the communication data chunks processed after the expert computation. Each stream executes in parallel, while execution within a stream is serial. The arrows between different streams in Fig.~\ref{fig17} represent event dependency. For example, the AllGather process of the communication data chunks in Stream 2 depends on the AllToAll process before the expert computation in Stream 1. The expert computation process depends on the completion of the D2D copy processes for all data chunks. The last two AllToAll processes in Stream 1 depend on the results of the expert computation.
\subsection{End-to-End time on Model}
In large model training, the initial context length is typically set to 4K or 8K, as in Llama3~\cite{abhimanyu2024llama3} which selects 8K. Here, an 8K sequence is chosen as input. With 2 nodes and 16 GPUs, $TP_Size=8$, $EP_Size=2$, and the MoE model selects 2x70B.

We test the Baseline against the $O_{1}/O_{2}/O_{3}$ optimizations, and the convergence curves are completely consistent, not affecting the numerical results. It can be seen that MoNTA selected the $O_{1}$ strategy.It can be observed that the performance of $O_{1}$ and $O_{2}$ is comparable and better than the Baseline, with the proportion of AllToAll communication reduced from 22\% to 10\%. The normalized end-to-end latency of the model is shown in Fig.~\ref{fig16}. $O_{1}$ and $O_{2}$ have similar performance, with an overall latency reduction of about 13\%. In contrast, the performance of $O_{3}$ declined compared to $O_{1}$ and $O_{2}$.
\begin{figure}[t]
\centering
\includegraphics[width=0.47\textwidth]{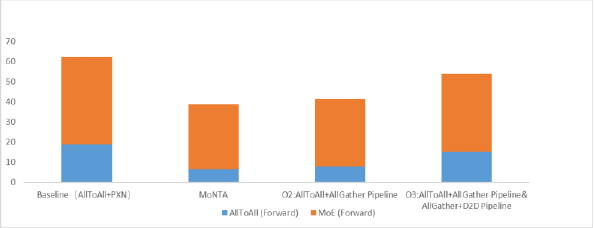} 
\caption{AllToAll Forward and MoE Forward.}
\label{fig15}
\end{figure}

\begin{figure}[t]
\centering
\includegraphics[width=0.47\textwidth]{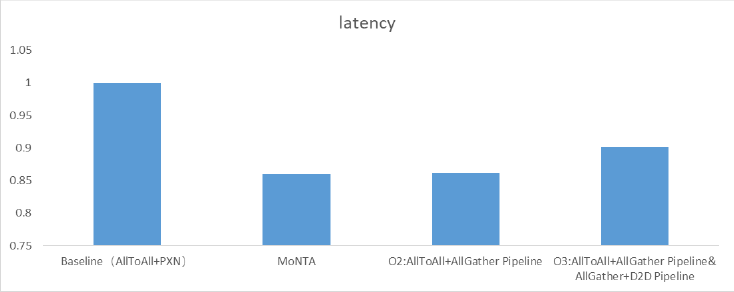} 
\caption{End-to-End model latency.}
\label{fig16}
\end{figure}

\subsection{Dissussions}
According to equations (8) and (9), we obtain the following latency for the parallel execution of AllToAll and AllGather:
\begin{equation}
\frac{AA_{j}}{AG_{j}}=\frac{(e-1)*B_{2}*r_{2}}{e*(t-1)*B_{1}*r_{1}}
\end{equation}
\noindent Where the communication bandwidth and efficiency for AllToAll are $B_{1}$ and $r_{1}$, respectively, and the communication bandwidth and efficiency for AllGather are $B_{2}$ and $r_{2}$.
For the 2x70B model with 2 nodes and 16 cards, $B_{2}/B_{1} =\frac{200}{25}=8,e=2,t=8$.
From Fig.~\ref{fig12}, we can obtain $r_{1}=0.741,r_{2}=0.803,r_{1}\approx r_{2},\frac{AA_{j}}{AG_{j}}=\frac{4}{7}$.
According to the performance model, the theoretical upper limit of the speedup ratio is:
\begin{equation}
\frac{T_{O_{3}}}{T_{base}}=\frac{7}{32}=0.21
\end{equation}

From Fig.~\ref{fig13}, it can be seen that when the input sequence length is 256K, the latency of $O_{3}$ compared to the Baseline is 0.253, approaching the theoretical limit, and the actual results are consistent with the theoretical analysis. The communication volume is related to microbatch size, sequence length, and hidden dimension; as the communication volume continues to increase, the optimization performance can be further enhanced.
When the input sequence length is 8K, we present a comparative analysis of the communication volume before and after optimization in Table 8:

\begin{table}[t]
\centering
\resizebox{0.47\textwidth}{!}{
\begin{tabular}{c|c|c|c|c}
\hline
\ Optimization strategy  & Communication primitives & Communication volume (M) & Communication efficiency & Theoretical Latency(ms)\\ \hline
Baseline & AllToAll & 256 & 74.1\%  & 6.909 \\ \hline
\multirow{2}{*}{$O_{1}$} & AllToAll & 32 & 63.2\%  & 1.012 \\
& AllGather & 256 & 77.6\%  & 1.443 \\ \hline
\multirow{3}{*}{$O_{2}/O_{3}(4 chunks)$} & AllToAll(per chunk) & 8 & 42.7\%  & 0.374 \\
& AllGather(per chunk) & 64 & 72.6\%  & 0.385 \\
& D2D copy(per chunk) & 64 & 80\%  & 0.05 \\
\hline
\end{tabular}
}
\caption{Comparison of communication volume before and after optimization.}
\end{table}

In the case of an 8K input sequence, $r_{1}=0.632,r_{2}=0.776$, under the $O_{1}$ strategy:
\begin{equation}
\frac{T_{O_{1}}}{T_{base}}=0.31
\end{equation}

Fig.~\ref{fig13} shows that $O_{1}/Baseline=0.30$, which is consistent with the theoretical estimate.
After splitting into 4 chunks for pipelining, the communication volume is reduced, and the AllToAll communication efficiency is around 40\%,$r_{1}=0.427,r_{2}=0.726,AA_{j}/AG_{j} =0.979$.The theoretical upper limit of speedup can be obtaine:
\begin{equation}
\frac{T_{O_{2}}}{T_{base}}=0.176
\end{equation}

\begin{equation}
\frac{T_{O_{3}}}{T_{base}}=0.164
\end{equation}

The theoretical time for AllToAll and AllGather for a single chunk is similar, allowing for better overlapping, as shown in (14) and (15). However, from Fig.~\ref{fig13}, it can be seen that when the input sequence length is 8K, the performance of $O_{2}/O_{3}$ is worse than that of $O_{1}$. This is because the AllGather communication kernel requires memory operations from NVLink to L3, leading to memory conflicts with the D2D copy kernel. At small data volumes, in addition to communication time, fixed latency dominates, lowering actual performance. Furthermore, inconsistencies in progress between processes result in misaligned dispatch times across different node hosts, affecting the execution time of the first chunk.
When the data volume is large, the AllGather communication time is long, but the memory copy efficiency is high, allowing for good hiding of the D2D copy. When the data volume is small, the scheduling of the AllGather and D2D copy kernels causes stalls, necessitating further optimization of kernel scheduling to reduce L3 memory access latency caused by scheduling for small data volumes, while also updating the performance model.
\section{Related Work}
\subsection{Mixture-of-Experts (MoE):} MoE has gradually gained popularity as an effective way to enhance the performance of large language models (LLMs), with its structure varying ~\cite{dmitry2020gshard}~\cite{albert2024mixtral}~\cite{damai2024deepseekmoe}. Gshard~\cite{dmitry2020gshard} was the first to introduce MoE into the Transformer model, demonstrating the significant potential of the MoE architecture in scaling model capacity. The Mixtral-8x7B~\cite{albert2024mixtral} model provides an approach to alleviate the issue of load imbalance, utilizing the dropless MoE algorithm proposed by Megablocks~\cite{trevor2022megablocks}. Megablocks addresses the problem of variable-length inputs with multiple experts using Grouped GEMM. DeepSeek MoE~\cite{damai2024deepseekmoe} offers a more fine-grained expert partitioning, allowing different experts to learn more specialized knowledge. Additionally, it introduces shared experts that activate for all input tokens.

\subsection{Optimization of all-to-all:}The PKU-DAIR laboratory proposed a high-performance distributed MoE training system, HetuMoE~\cite{xiaonan2022hetumoe}, which supports various gate operator optimizations, such as Top1 and K Top1. It developed a hierarchical communication operator, Hierarchical AllToAll, for single NIC network nodes. Tutel~\cite{changho2021tutel} implements adaptive parallelism switching, dynamically adjusting the combination of data parallelism (DP), model parallelism (MP), and expert parallelism (EP) based on the characteristics of the input data. It segments input tokens to form a pipeline, allowing expert computation and All-to-All communication to be pipelined. Flux~\cite{liwen2024flux} proposed overlapping tensor parallel computation and communication through kernel fusion methods. FastMoE~\cite{jiaao2001fastmoe} employs a dynamic routing algorithm that selects experts for computation based on their load and the characteristics of the input data. PipeMoE~\cite{shaohuai2023pipemoe} introduced dispatch, expert computation, and combine pipelining, showing improved performance compared to FasterMoE and Tutel. MPMoE~\cite{zheng2024mpmoe} proposed profile-based algorithm optimizations for pipelining and memory reuse. DeepSpeed-TED~\cite{siddharth2023deepseepted} integrates Zero data parallelism, Megatron-LM's~\cite{mohammad2020megatronlm}~\cite{deepak2021megatron}~\cite{vijay2022reducing} tensor parallelism, and DeepSpeed-MOE's expert parallelism, now incorporated into DeepSpeed. It reduces All-to-All communication volume through Duplicate Token Dropping. However, optimizations for the MoE module under tensor parallelism have not been merged into the official DeepSpeed, and we submitted patch1. Additionally, leveraging cluster communication across different hardware resources, we achieved pipelining for AllToAll and AllGather, realizing a network-traffic-aware parallel optimization scheme.

\section{Conclusion and Future Works}
This paper proposes a network-traffic-aware parallel optimization method called MoNTA, which utilizes inter-node and intra-node communication resources to achieve pipelining for inter-node AllToAll and intra-node communication. It establishes a performance model for communication volume, communication efficiency, and parallel schemes, effectively overlapping MoE AllToAll communication. Additionally, it analyzes communication conflict issues during the training process of MoE models and presents a communication priority scheme. Finally, it proposes a distributed parallel training extension method for the long context MoE models. Experimental results show that MoNTA can achieve a performance improvement of approximately 8x in AllToAll communication under 8-card tensor parallelism compared to the baseline. When using 16 A800 cards to train a 2x70B model with an 8K sequence, the overall latency performance is improved by 13\% compared to the baseline.

The next step will be analyzing the impact of kernel scheduling on MoE parallel optimization performance, refining the performance model for training and inference, and integrating it into the framework and operator implementations. Similar to Flux~\cite{liwen2024flux}, achieving overlapping of AllToAll and expert computation through software kernel fusion is also a direction for future exploration.

\appendix

\bigskip

\bibliography{aaai25}

\begin{thebibliography}{19}
\providecommand{\natexlab}[1]{#1}

\bibitem[{Abhimanyu et~al.(2024)Abhimanyu, Abhinav, Abhinav, Abhishek, Ahmad, and Aiesha}]{abhimanyu2024llama3}
Abhimanyu, D.; Abhinav, J.; Abhinav, P.; Abhishek, K.; Ahmad, A.-D.; and Aiesha, L. 2024.
\newblock The Llama 3 Herd of Models.
\newblock \emph{arXiv preprint arXiv:2407.21783}.

\bibitem[{Albert~Q. et~al.(2024)Albert~Q., Alexandre, Antoine, Arthur, and Blanche}]{albert2024mixtral}
Albert~Q., J.; Alexandre, S.; Antoine, R.; Arthur, M.; and Blanche, S. 2024.
\newblock Mixtral of Experts.
\newblock \emph{arXiv preprint arXiv:2401.04088}.

\bibitem[{Changho et~al.(2021)Changho, Wei, Yifan, Ziyue, Ze, Han, Zilong, Rafael, Jithin, Prabhat, Joe, Peng, Fan, Mao, and Yongqiang}]{changho2021tutel}
Changho, H.; Wei, C.; Yifan, X.; Ziyue, Y.; Ze, L.; Han, H.; Zilong, W.; Rafael, S.; Jithin, J.; Prabhat, R.; Joe, C.; Peng, C.; Fan, Y.; Mao, Y.; and Yongqiang, X. 2021.
\newblock Tutel: Adaptive mixture-of-experts at scale.
\newblock \emph{arXiv preprint arXiv:2206.03382}.

\bibitem[{Damai et~al.(2024)Damai, Chengqi, Chenggang, R.X., Huazuo, Deli, Jiashi, Wangding, Xingkai, Y., Zhenda, Y.K., Panpan, Fuli, Chong, Zhifang, and Wenfeng}]{damai2024deepseekmoe}
Damai, D.; Chengqi, D.; Chenggang, Z.; R.X., X.; Huazuo, G.; Deli, C.; Jiashi, L.; Wangding, Z.; Xingkai, Y.; Y., W.; Zhenda, X.; Y.K., L.; Panpan, H.; Fuli, L.; Chong, R.; Zhifang, S.; and Wenfeng, L. 2024.
\newblock DeepSeekMoE: Towards Ultimate Expert Specialization in Mixture-of-Experts Language Models.
\newblock \emph{arXiv preprint arXiv:2401.06066}.

\bibitem[{Deepak et~al.(2021)Deepak, Mohammad, Jared, Patrick, and Mostofa}]{deepak2021megatron}
Deepak, N.; Mohammad, S.; Jared, C.; Patrick, L.; and Mostofa, P. 2021.
\newblock Eﬀicient Large-Scale Language Model Training on GPU Clusters Using Megatron-LM.
\newblock \emph{arXiv preprint arXiv:2104.04473}.

\bibitem[{Dmitry et~al.(2020)Dmitry, HyoukJoong, Yuanzhong, Dehao, Orhan, Yanping, Maxim, Noam, and Zhifeng}]{dmitry2020gshard}
Dmitry, L.; HyoukJoong, L.; Yuanzhong, X.; Dehao, C.; Orhan, F.; Yanping, H.; Maxim, K.; Noam, S.; and Zhifeng, C. 2020.
\newblock GShard: Scaling giant models with conditional computation and automatic sharding.
\newblock \emph{arXiv preprint arXiv:2006.16668}.

\bibitem[{Hao, Matei, and Pieter(2023)}]{hao2023ring}
Hao, L.; Matei, Z.; and Pieter, A. 2023.
\newblock Ring Attention with Blockwise Transformers for Near-Infinite Context.
\newblock \emph{arXiv preprint arXiv:2310.01889}.

\bibitem[{Jiaao et~al.(2022)Jiaao, Jidong, Tiago, Haojie, Fuwen, Shangfeng, and Qin}]{jiaao2001fastmoe}
Jiaao, H.; Jidong, Z.; Tiago, A.; Haojie, W.; Fuwen, L.; Shangfeng, S.; and Qin, L. 2022.
\newblock FasterMoE: Modeling and optimizing training of large-scale dynamic pre-trained models.
\newblock In \emph{Proceedings of the 27th ACM SIGPLAN Symposium on Principles and Practice of Parallel Programming}, 120--134.

\bibitem[{Li-Wen et~al.(2024)Li-Wen, Wenlei, Qi, Chengquan, Ningxin, Yinmin, Xuanrun, Zuquan, Chengji, Ziheng, Haibin, Xin, and Xin}]{liwen2024flux}
Li-Wen, C.; Wenlei, B.; Qi, H.; Chengquan, J.; Ningxin, Z.; Yinmin, Z.; Xuanrun, Z.; Zuquan, S.; Chengji, Y.; Ziheng, J.; Haibin, L.; Xin, J.; and Xin, L. 2024.
\newblock FLUX: Fast Software-based Communication Overlap On GPUs Through Kernel Fusion.
\newblock \emph{arXiv preprint arXiv:2406.06858}.

\bibitem[{Liliang et~al.(2024)Liliang, Yang, Yadong, Yelong, Chen, and Weizhu}]{liliang2024samba}
Liliang, R.; Yang, L.; Yadong, L.; Yelong, S.; Chen, L.; and Weizhu, C. 2024.
\newblock Samba: Simple Hybrid State Space Models for Efficient Unlimited Context Language Modelingt.
\newblock \emph{arXiv preprint arXiv:2406.07522}.

\bibitem[{Mohammad~Shoeybi(2020)}]{mohammad2020megatronlm}
Mohammad~Shoeybi, R. P. P. L. J. C. B.~C., Mostofa~Patwary. 2020.
\newblock Megatron-LM: Training Multi-Billion Parameter Language Models Using Model Parallelism.
\newblock \emph{arXiv preprint arXiv:1909.08053}.

\bibitem[{Opher et~al.(2024)Opher, Barak, Hofit, Gal, Jhonathan, Itay, Erez, Shaked, Yonatan, Shai, Omri, Raz, Tomer, Amir, Roman, Michael, Avashalom, Nir, Noam, Erez, Mor, and Yoav}]{opher2024jamba}
Opher, L.; Barak, L.; Hofit, B.; Gal, C.; Jhonathan, O.; Itay, D.; Erez, S.; Shaked, M.; Yonatan, B.; Shai, S.-S.; Omri, A.; Raz, A.; Tomer, A.; Amir, B.; Roman, G.; Michael, G.; Avashalom, M.; Nir, R.; Noam, R.; Erez, S.; Mor, Z.; and Yoav, S. 2024.
\newblock Jamba: A Hybrid Transformer-Mamba Language Model.
\newblock \emph{arXiv preprint arXiv:2403.19887}.

\bibitem[{Shaohuai et~al.(2023)Shaohuai, Xinglin, Xiaowen, and Bo}]{shaohuai2023pipemoe}
Shaohuai, S.; Xinglin, P.; Xiaowen, C.; and Bo, L. 2023.
\newblock PipeMoE: Accelerating Mixture-of-Experts through Adaptive Pipelining.
\newblock In \emph{IEEE Conference on Computer Communications}.

\bibitem[{Shriram, Rudolf, and Richard(2001)}]{shriram2001connection}
Shriram, S.; Rudolf, R.; and Richard, B. 2001.
\newblock Connection-level analysis and modeling of network traffc.
\newblock In \emph{Proceedings of the 1st ACM SIGCOMM Workshop on Internet measurement}, 99--103.

\bibitem[{Siddharth et~al.(2023)Siddharth, Olatunji, Ammar, Samyam, Yuxiong, and Abhinav}]{siddharth2023deepseepted}
Siddharth, S.; Olatunji, R.; Ammar, A.~A.; Samyam, R.; Yuxiong, H.; and Abhinav, B. 2023.
\newblock A Hybrid Tensor-Expert-Data Parallelism Approach to Optimize Mixture-of-Experts Training.
\newblock \emph{arXiv preprint arXiv:2303.06318}.

\bibitem[{Trevor et~al.(2022)Trevor, Deepak, Cliff, and Matei}]{trevor2022megablocks}
Trevor, G.; Deepak, N.; Cliff, Y.; and Matei, Z. 2022.
\newblock Megablocks: Efficient Sparse Training With Mixture-of-Experts.
\newblock \emph{arXiv preprint arXiv:2211.15841}.

\bibitem[{Vijay et~al.(2022)Vijay, Jared, Sangkug, Lawrence, Michael, Mohammad, and Bryan}]{vijay2022reducing}
Vijay, K.; Jared, C.; Sangkug, L.; Lawrence, M.; Michael, A.; Mohammad, S.; and Bryan, C. 2022.
\newblock Reducing Activation Recomputation in Large Transformer Models.
\newblock \emph{arXiv preprint arXiv:2205.05198}.

\bibitem[{Xiaonan et~al.(2022)Xiaonan, Pinxue, Xupeng, Tong, and Bin}]{xiaonan2022hetumoe}
Xiaonan, N.; Pinxue, Z.; Xupeng, M.; Tong, Z.; and Bin, C. 2022.
\newblock HetuMoE: An Efficient Trillion-scale Mixture-of-Expert Distributed Training System.
\newblock \emph{arXiv preprint arXiv:2203.14685}.

\bibitem[{Zheng et~al.(2024)Zheng, Yaqi, Hulin, Donglin, Chuang, Xiaobo, and Dazhao}]{zheng2024mpmoe}
Zheng, Z.; Yaqi, X.; Hulin, W.; Donglin, Y.; Chuang, H.; Xiaobo, Z.; and Dazhao, C. 2024.
\newblock MPMoE: Memory Efficient MoE for Pre-Trained Models With Adaptive Pipeline Parallelism.
\newblock \emph{IEEE Transactions on Parallel and Distributes Systems.}, 35(6): 998--1011.

\end{thebibliography}

\end{document}